\documentclass{article} 
\usepackage{hyperref}
\usepackage{url}
\usepackage{verbatim}

\usepackage[top=1.2in, bottom=1.2in, left=1.2in, right=1.2in]{geometry}

\usepackage{amsthm,amsmath,amssymb}
\usepackage{latexsym}
\usepackage{graphicx}
\usepackage{subfigure}
\usepackage{algorithm}
\usepackage{algpseudocode} 

\usepackage{color}

\newcommand{\C}{\mathcal{C}}
\newcommand{\pp}{\mathcal{P}}

\title{An Incremental Reseeding Strategy for Clustering}
\date{}

\author{
Xavier Bresson\thanks{University of Lausanne, ({\tt xavier.bresson@unil.ch})} \and
Huiyi Hu\thanks{Department of Mathematics, University of California,  ({\tt huiyihu@math.ucla.edu})} \and 
Thomas Laurent\thanks{Department of Mathematics, Loyola Marymount University ({\tt  tlaurent@lmu.edu})} \and
Arthur Szlam\thanks{Department of Mathematics, The City College of New York ({\tt  aszlam@ccny.cuny.edu})} \and
James von Brecht\thanks{Department of Mathematics, University of California Los Angeles ({\tt jub@math.ucla.edu})}  
}

\begin{document}

\maketitle

\begin{abstract}
 In this work we propose a simple and easily parallelizable algorithm for multiway graph partitioning. The algorithm alternates between three basic components: diffusing seed vertices over the graph, thresholding the diffused seeds, and then randomly reseeding the thresholded clusters. We demonstrate experimentally that the proper combination of these ingredients leads to an algorithm that achieves state-of-the-art performance in terms of cluster purity on standard benchmarks datasets. Moreover, the algorithm runs an order of magnitude faster than the other algorithms that achieve comparable results in terms of accuracy \cite{yang2012clustering}. We also describe a coarsen, cluster and refine approach similar to  \cite{GRACLUS2007,metis98} that removes an additional order of magnitude from the runtime of our algorithm while still maintaining competitive accuracy. 
\end{abstract}

\section{Introduction}
One of the most basic unsupervised learning tasks is to automatically partition data into clusters based on similarity. A standard scenario is that the data is represented as a weighted graph, whose data points correspond to vertices on the graph and whose edges encode the similarity between data points. Many of the most popular and widely used clustering algorithms, such as spectral clustering, fall into this category. Despite the vast literature on graph-based clustering, the field remains an active area for both theoretical and practical research.  

In this work, we propose a resampling-based spectral algorithm for multiway graph partitioning that achieves a good combination of accuracy and efficiency on graphs that contain reasonably well-balanced clusters of medium scale. The algorithm is simple, intuitive, and easy-to-implement. It also parallelizes trivially, and can therefore scale gracefully to large numbers of clusters as well as large numbers of graph vertices. We demonstrate experimentally that the algorithm achieves state-of-the-art performance in terms of cluster purity on standard benchmarks, while running an order of magnitude faster than the other highly accurate clustering methods, e.g. \cite{yang2012clustering}. The appeal of our algorithm arises from the combination of simplicity, accuracy, and efficiency that it provides. 

The straightforward implementation of our algorithm (in serial) runs two orders of magnitude slower than popular multiscale coarsen-and-refine algorithms, such as \cite{GRACLUS2007,metis98}.  We show experimentally that a similar combination of coarsening and refinement can remove an order of magnitude from the runtime of our algorithm while maintaining a level of accuracy between state-of-the-art (but expensive) direct approaches \cite{yang2012clustering} and heavily optimized multigrid ones \cite{GRACLUS2007,metis98}.\\

\section{Description of the Algorithm}   
\begin{figure}[t!]
\centering
\includegraphics[width=12cm]{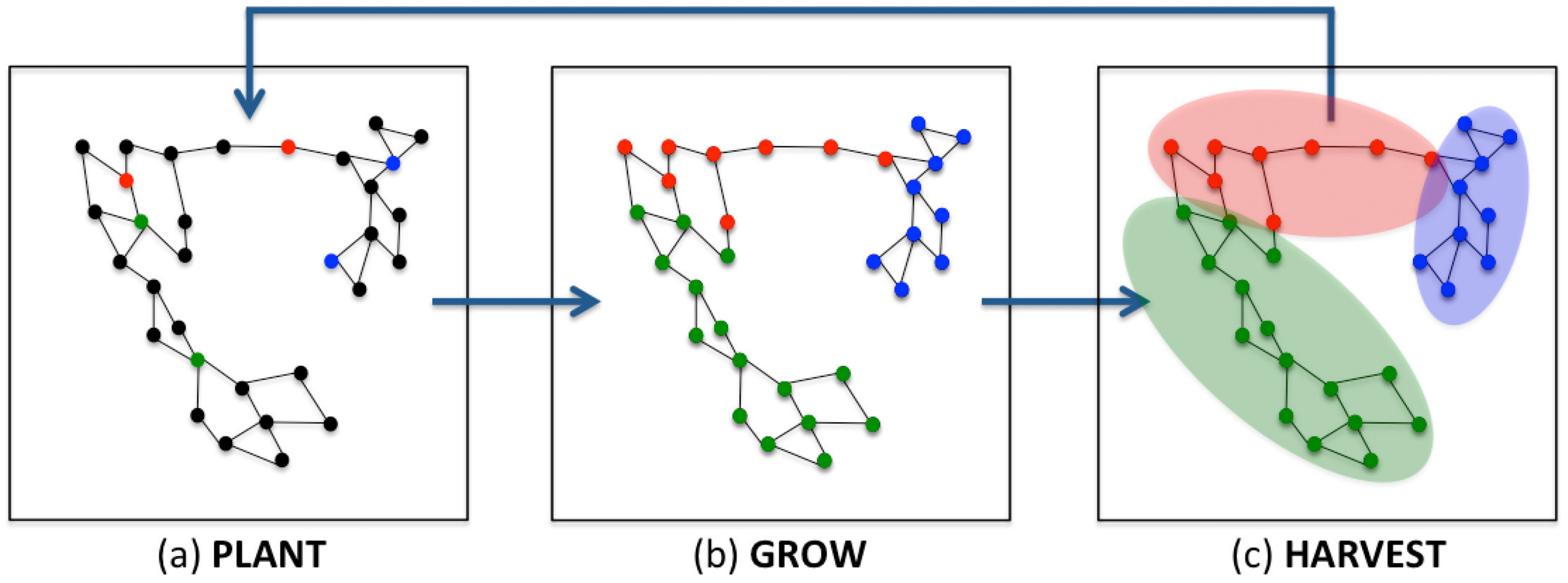}
\caption{Illustration of the Incremental Reseeding (INCRES) algorithm for $R=3$ clusters. The colors red, blue, and green are used to identify the clusters. Figure (a): At this stage of the algorithm, $m=2$ seeds are randomly planted in the clusters computed from the previous iteration. Figure (b): The seeds grow with the random walk operator. Figure (c): A new partition of the graph is obtained and it will be used to plant $m+\Delta_m$ seeds into these clusters at the next iteration.}
\label{figIllustration}
\end{figure}

A main idea behind our algorithm arises from a well-known (and widely used) property of the random walk on a graph. Specifically, a random walker started in a low conductance cluster is unlikely to leave the cluster quickly. This fact provides the basis for transductive methods such as \cite{Zhu03semi-supervisedlearning}, as well as for ``local'' clustering methods such as \cite{Nibble2013}. Each of these works require an initial guess for the location of clusters.  In the transductive case, this location information comes in the form of class labels provided by an oracle. In the case of \cite{Nibble2013}, the ``label'' information comes in the form of an \emph{a-priori} assumption of smallness or locality (i.e. a small random-walk extent) for the cluster that contains a specified seed vertex. A partitioning of the whole graph is then obtained from these local clusterings via a sequential extraction of small clusters with random choices for the single ``label'' or seed vertex  \cite{Nibble2013}.   

Our algorithm combines ideas from these approaches. Assume that the graph has $R$ well-defined clusters of comparable size and low conductance. If we could assign to each cluster an initial vertex, then we might expect good results from a transductive label propagation by using these initial assignments as labels. In an unsupervised context we cannot, of course, place a seed in each cluster as we do not know the clusters themselves beforehand. To overcome this, we instead place a handful of seeds at random. We then perform a few steps of a random walk using the selected vertices as initial positions. We obtain a temporary clustering by assigning each node on the graph to its nearest seed, and then reseed the labels from these temporary clusters. If the clusters improve then the seeds will likely improve, and vice-versa. This incites a feed-back loop and we get a virtuous cycle. We can then excite the speed and improve the quality of this cycle by gradually drawing more and more seeds throughout the process. We refer to this idea as an \emph{incremental reseeding strategy}, and we depict this cyclic process graphically in figure \ref{figIllustration}.  

%

\subsection{Implementation Details and Practical Improvements}
To formalize these ideas, let $G = (V,W)$ denote a weighted, undirected graph on $N$ vertices  $V = \{v_1,\ldots,v_N\}$ with symmetric edge weights $W = \{w_{ij} \}^{N}_{i,j=1}$ that encode a measure of similarity between each pair $(i,j)$ of vertices. Let $D$ denote the diagonal matrix of (weighted) vertex degrees. We propose the following randomized, iterative algorithm for partitioning such a graph into $R$ classes. First, generate an initial partition $\pp^{0}=(\C^0_1, \ldots, \C^0_R)$ of the graph $V = \C^{0}_{1} \; \cup \ldots \cup \; \C^{0}_{R}$ into $R$ disjoint clusters $\C^0_r$ by assigning each vertex $v_i$ to one of the $R$ classes uniformly at random. Let $m=1$ denote the initial number of seeds. At each of the successive iteration, we update the current partition $\pp^{k} =(\C^k_1, \ldots, \C^k_R)$ according to the steps outlined in algorithm \ref{main} below. We refer to algorithm \ref{main} as the Incremental Reseeding algorithm (INCRES).

\begin{algorithm}[H]
\caption{ INCRES Algorithm} \label{main}

\begin{algorithmic}[]
\State
{\bf Input:} Similarity matrix $W$, seed increment  $\Delta_{m}$, number of clusters $R$. 
\State {\bf Initialization:} $m=1$, random partition $\pp = (\C_1, \ldots, \C_R)$
\Repeat{}
\State $F  = \text{PLANT}(\pp, m)$   
\State $F  \gets \text{GROW}(F,W)$  
\State $\pp \gets \text{HARVEST}(F)$   
\State{$m \gets m+\Delta_m$ } 
\Until{  $\pp$ converges }
\State{{\bf Output:} $\pp$ }
\end{algorithmic}
\end{algorithm}

A variety of different possibilities exist for the choices of the PLANT, GROW, and HARVEST subroutines used in this basic framework. We discuss the basic choices we use in our experiments, as well as a few variants that we have found prove useful in certain special circumstances. The first routine, PLANT, implements the basic reseeding strategy:

\begin{algorithm}[H]
\label{PLANT}
\begin{algorithmic}[]
 \Function{PLANT}{$\pp,m$}
\State Initialize $F$ as an $N$-by-$R$ sparse matrix of zeros. 
 \For{$r=1$ to  $R$}
 \State Select a subset $V_r$ of $\C_r$ with $m$ vertices by sampling uniformly without replacement.
 \State Set the $r^{ {\rm th} }$ column of $F$ equal to the indicator function $\mathbf{1}_{V_r}$ of $V_r$.
\EndFor
 \State \textbf{return}  $F$.
 \EndFunction
\end{algorithmic}
\end{algorithm} 

We use this outline as the basis of our implementation of the PLANT routine used in our experiments. If the number of seeds $m$ happens to exceed the size of one of the clusters $\C_r$ at any given iteration, we simply reinitialize $m$ as the size of the smallest cluster $\C_{r}$ at that iteration. We then return this value from PLANT and increment $m \gets m + \Delta_{m}$ as before. The overall computational cost of the PLANT function proves modest. The main computational burden lies identifying the clusters and in generating the random sample.

 \begin{figure*}[b]
  \begin{minipage}{.49\linewidth}
  \begin{algorithm}[H]
\label{GROW}
\begin{algorithmic}[]
 \Function{GROW}{$F,W$}
\While{ $\min_i \min_r  F_{i,r} = 0$ }
\State $F \gets \left( WD^{-1} \right) F$
\EndWhile  
 \State \textbf{return}  $F$.
 \EndFunction
\end{algorithmic}
\end{algorithm} 
\end{minipage}
\hfill 
\begin{minipage}{.49\linewidth}
\begin{algorithm}[H]
\label{HARVEST}
\begin{algorithmic}[]
  \Function{HARVEST}{$F$}
 \For{$r=1$ to  $R$}
 \State $\C_r=\{i : F_{i,r} \ge F_{i,s} \forall  s \neq r \}$
 \EndFor
 \State \textbf{return}  $\pp=(\C_1, \ldots, \C_R)$
 \EndFunction
\end{algorithmic}
\end{algorithm} 
\end{minipage}
 \end{figure*}

The simplest choices for the GROW and HARVEST functions appear below. We use this particular implementation of the GROW routine in all of our experiments, although we have experimented with a number of different choices as well. In particular, we have found that replacing the random walk step $F \gets WD^{-1}$ with a diffusion step $F \gets D^{-1}W$ will give similar results in many circumstances. Occasionally, we have found that utilizing a ``personalized Page-Rank'' step
$$F \gets \alpha W D^{-1} F + (1-\alpha) F_{0} $$
can give better performance on small data sets that contain a large (relative to the size of the data set) number of clusters. Here $0 < \alpha < 1$ denotes the random walk extent (usually set to $\alpha = .85$) and $F_{0}$ denotes the indicator matrix that initializes the GROW routine. A step of this form amounts to measuring similarity between vertices in the same manner used in either  \cite{NibblePageRank} or \cite{yang2012clustering}, up to replacing $D^{-1}W$ with $D^{-1/2}WD^{-1/2}$ in the latter case. By-and-large the INCRES algorithm proves robust to the particular implementation of GROW, so long as it realizes the basic idea of label propagation in one form or another. In any case, the main computational burden of the algorithm arises from the GROW routine. The procedure will terminate once the labels produced by PLANT have propagated throughout the entire graph. This requires a connected graph and a computational cost of $O(R E\mathrm{diam}(G))$ in the worst case, where $E$ denotes the number of edges in the graph and $\mathrm{diam}(G)$ denotes its diameter.  We can, however, introduce an ``economy'' version of GROW that allows us to handle datasets with a large number of clusters $R$ without having to store a full matrix $F$ of indicators for each cluster. We also use this implementation of HARVEST for all of our experiments, and we have yet to run across a situation in which modifying it would prove useful. Its cost provides only a negligible contribution to the overall cost of each step of the algorithm. As our experiments will show, this simple combination of ingredients (and in particular the PLANT function) turns the heuristic outlined above into a reliable clustering algorithm.

\subsection{Relation with Other Work}
As we previously discussed, our method relies upon and incorporates number of ideas from transductive learning. In particular, we leverage the notion of label propagation \cite{Zhu03semi-supervisedlearning}.  In the standard label propagation framework, an oracle provides a set of labeled points or vertices. These labeled points form either nonzero initial conditions or heat sources for a discrete heat equation on the graph. The second step of the algorithm outlined above (which we term GROW below) precisely corresponds with a label propagation of the random labels returned from the first step of the algorithm (which we term PLANT below). The ${\rm NIBBLE}$ algorithm and its relatives \cite{lovasz1993random,Nibble2004,NibblePageRank,Nibble2013} use a similar idea to get an unsupervised clustering method from label propagation by planting random seeds. These works cluster the entire graph in a sequential manner, and each cluster in the sequence is obtained from a localized cluster around a single random vertex. We perform multiway partitioning directly, instead of recursively, and utilize a significantly different random seeding strategy. Our algorithm also alternates between label propagation (in step 2) and thresholding (in step 3). The idea of iteratively alternating between a few steps of label propagation and subsequent thresholding has also appeared in a transductive learning context \cite{MBOPAMI}, although the presence of labelled information results in a different implementation of the propagation step. The non-negative matrix factorization method \cite{yang2012clustering} also incorporates random walk information in a manner that resembles step 2, but otherwise the underlying principles of the algorithms differ.

The algorithms GRACLUS \cite{GRACLUS2007} and METIS \cite{metis98} directly inspired the multigrid version of our algorithm.  We use the same coarsening algorithm, but rely upon a different clustering on the coarsest scale (algorithm \ref{main} vs. kernelized $k$-means or pure spectral clustering) and we use a different refinement technique.  Algorithm \ref{main} relates to the kernelized $k$-means procedure used in GRACLUS even in the single level case: we can essentially interpret the GROW function (step 2) as the ``maximization'' step in an alternating minimization for a kernelized $k$-means. Here the kernel is a power of the normalized weights and the power may depend on the cluster, so it is not exactly the same. The ``expectation'' step in our algorithm is replaced by sampling, and instead of having a single representative for a class, the number of representatives increases as the algorithm progresses. Using power iterations of the weight matrix $W$ directly for clustering has appeared in \cite{power_clustering,diffusion_maps_pami}. These works utilize the power iterations to generate an embedding of the vertices of the graph, which is then clustered using $k$-means. These methods can also be considered as kernelized $k$-means methods, with a power of the weights providing the kernel.

Because the GROW function we use iterates the random walk on the graph, our algorithm is a form of spectral clustering.  However,  our main contribution to the clustering problem, and the primary novelty in our algorithm, is the {\it incremental reseeding process}. This process is not specific to the GROW function presented here--- it seems to be quite universal and can be adapted to other clustering methods. However, combining reseeding with the random walk method offers an excellent combination of accuracy, speed, and robustness.

\subsection{A Multigrid Speedup}
As we just discussed, the main computational cost in algorithm \ref{main} stems from the multiplication of $F$ by the random walk matrix. Much of this multiplication is wasted if the graph has a large number of vertices and a relatively small number of high quality clusters; a typical random walker would take a long time to reach the boundary of the cluster in which it starts. A standard approach for dealing with this difficulty is to coarsen the graph, solve the clustering problem on a coarsened graph and then successively refine the clustering to transfer back to the original graph \cite{GRACLUS2007,metis98}.  

We follow the same coarsening procedure \cite{GRACLUS2007,metis98} in our multilevel approach. We begin with each vertex in the graph unmarked. We then pass through the vertices and associate each vertex to its most similar neighbor, then mark the current vertex and its neighbors. If a vertex has no unmarked neighbors then it remains a singleton. The coarsened weights are just the sum of all the weights in each mini-cluster. That is, if the new vertex $\overline{v}_k=\{v_{k_1}, v_{k_2}\}$ then 
\[\overline{W}_{kj}= W_{k_1 j_1}+  W_{k_1 j_2}+W_{k_2 j_1} +W_{k_2 j_2}.\]  
Our experiments also show that we can obtain accuracies competitive with \cite{GRACLUS2007,metis98} on benchmarks like 20NEWS and MNIST with even a trivial refinement procedure: we assign each element in a coarsened node the class label of its parent.  However, we can achieve higher accuracy with a more careful refinement: to get from each scale to the next finer scale, we run a slightly modified version algorithm \ref{main} initialized from the clustering at the coarser scale. This modification allows the random walk to cover the graph much faster than starting from one seed per cluster.

Our refinement procedure proceeds as follows. Let $N_{ {\rm sm} }$ denote the size of the coarsest graph returned by the coarsening procedure and let $L$ denote the corresponding number of levels in the hierarchy. We initialize our refinement procedure by first computing a base clustering of the coarsest graph by performing the INCRES algorithm at the coarsest level for a fixed number $k_{{\rm sm }}$ of iterations. This procedure returns a number $m_{ {\rm sm} }$ of seeds upon termination. We then let
$$
\alpha_{ {\rm seed} } := \left( \frac{N}{N_{ {\rm sm} } } \right)^{ \frac1{L - 1} } \qquad \alpha_{ {\rm iter} } := \left(\frac{ k_{{\rm sm }} }{2} \right)^{ \frac1{L - 1} }
$$    
denote the amount by which we will increase the number of seeds and decrease the number of iterations at each level, respectively. In other words, if level $l=1$ denotes the coarsest level we let $m_{1} = m_{ {\rm sm} }$ and $k_{1} = k_{ {\rm sm}}$ initially. For levels $2 \leq l \leq L$ we draw $m_{l} := \alpha_{ {\rm seed} } m_{l-1}$ seeds at each PLANT step and perform a total of $k_{l} :=  k_{l-1} /  \alpha_{ {\rm iter} }$ iterations of the INCRES algorithm. Note that with these choices we have
$$
\frac{ m_{L} }{ N } = \frac{ m_{1} }{N_{ {\rm sm} } } \qquad \text{and} \qquad k_{L} = 2.
$$ 
In other words, at each PLANT step we draw approximately the same proportion of seeds at every level in the hierarchy. We also geometrically decrease the total number of multiplications required at each level. In this way, the parameters $k_{ {\rm sm} }$ and $N_{ {\rm sm} }$ of the initial clustering at the coarsest level completely determine the dynamics of the refinement procedure.    
  
 



\section{Experiments}

We compare our method against four clustering algorithms that rely on variety of different principles. We select algorithms that, like our algorithm, partition the graph in a direct, non-recursive manner. The NCut  algorithm \cite{YS2003} is a widely used spectral algorithm that relies on a post-processing of the eigenvectors of the graph Laplacian to optimize the normalized cut energy. The NMFR algorithm  \cite{yang2012clustering} uses non-negative matrix factorization and graph-based random walk principles in order to factorize and regularize the original input similarity matrix. The LSD algorithm  \cite{arora2011clustering} provides another non-negative matrix factorization algorithm. It aims at finding a left-stochastic  decomposition of the similarity matrix. The MTV algorithm from \cite{BLUV13} provides a total-variation based algorithm that attempts to find an optimal multiway Cheeger cut of the graph by using $\ell^1$ optimization techniques. The last three algorithms (NMFR, LSD and MTV)
 all use NCut in order to obtain an initial partition. By contrast, we initialize our algorithm with a random partition. We use the code available from \cite{YS2003} for NCut, the code available from \cite{yang2012clustering} to test the two non-negative matrix factorization algorithms (NMFR and LSD) and the code available  from  \cite{BLUV13} for the MTV algorithm.
 
{\bf The Seed Increment Parameter $\Delta_m$:}  Recall that $\Delta_m$ denotes the amount by which we increase the number of seeds $m$ sampled from each class during an iteration of the algorithm. A larger value of $\Delta_m$ will quickly increase the seed number $m$ and the algorithm therefore converges more quickly. On the other hand, a small value of $\Delta_m$ will allow the algorithm to progress more slowly and allow the algorithm more freedom in its random exploration of the possible partitions of the graph. This often results in higher-quality clusterings. The choice of $\Delta_m$ should therefore reflect a good compromise between speed and quality. In practice, we generally select
 $$
 \Delta_m=\text{speed} \times 10^{-4} \times \frac{N}{R}
 $$
 with a proportionality constant {\bf speed} between 1 and 10. In the experiments we show results for $\text{{\bf speed}}=5$ (the default setting of our algorithm) and  $\text{\bf{speed}}=1$ (for a slower but more accurate algorithm).

 {\bf The Datasets:}  We provide experimental results on five text datasets (20NEWS, CADE, RCV1, WEBKB4, CITESEER) and four handwritten digits image datasets (MNIST, PENDIGITS, USPS, OPTDIGITS). We processed the text data sets by removing a list of stop words as well as by removing all words with fewer than twenty occurrences (for 20NEWS) and fewer than five occurrences (for all others) across the corpus. We then construct a 5-NN graph based on the cosine similarity between tf-idf features. For variety, we include some weighted graphs (RCV1 and CITESEER) as well as some unweighted graphs (20NEWS, CADE and WEBKB4). For MNIST, PENDIGITS and OPTDIGITS  we use the similarity matrices constructed by \cite{yang2012clustering}, where the authors first extract scattering features \cite{bruna2013invariant} for images before calculating an unweighted 10-NN graph. For USPS we constructed a weighted 10-NN graph from the raw data without any preprocessing. We provide the source for these data sets and more details on their construction in the appendix.



\begin{figure}[h!]
\centering
\subfigure[20NEWS (first 3mn)]{\includegraphics[width=6.8cm]{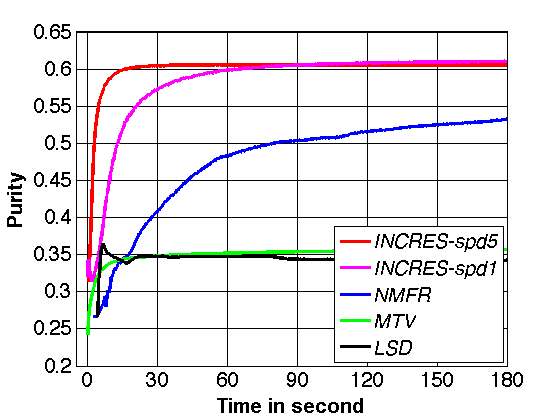}}
\subfigure[20NEWS (first 20mn)]{\includegraphics[width=6.8cm]{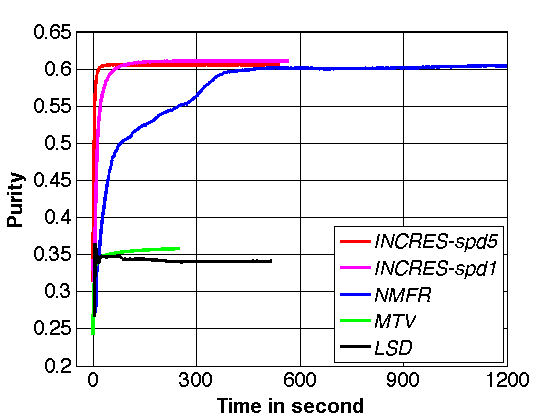}} \\
\subfigure[MNIST (first 3mn)]{\includegraphics[width=6.8cm]{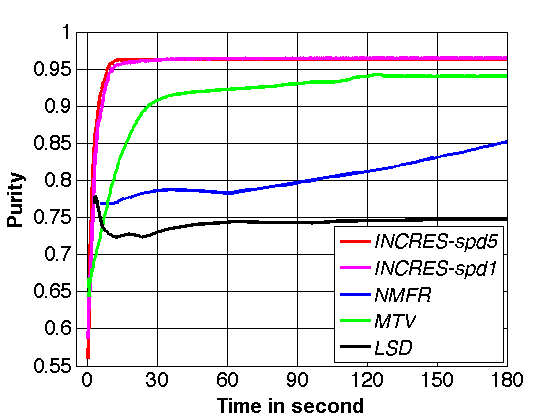}}
\subfigure[MNIST (first 20mn)]{\includegraphics[width=6.8cm]{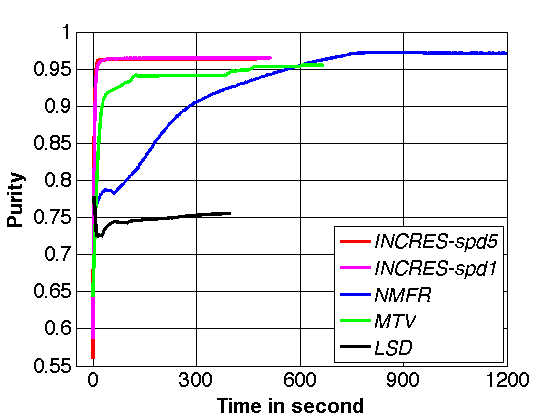}} \\  
\caption{ Purity curves for the four algorithms considered on two classical data sets (20NEWS and MNIST). We plot purity against time for each algorithm over two different time windows.}
\label{prtvstime}
\end{figure}

{\small
\begin{table}[h]
\centering
\begin{tabular}{c|c|c|c|c|c|c|c|c|c}
 &   {\bf LSD}   &{\bf NMFR}& {\bf MTV} & {\bf INCRES}& {\bf INCRES} \\
                                            &                    &                    &                  &         {\it (speed 1)}                          &           {\it (speed 5)      }                  \\  
\hline	
20NEWS -- time to reach 60\% purity  & --   &        469s &   --   & 62.8s  & 16.7s  \\
MNIST -- time to  reach 95\% purity  & -- &  566s &  458s & 10.3s &  8.8s  \\
\end{tabular}
\label{tab:crosstime}
\caption{Speed/accuracy trade-off: computational time required for each algorithm to reach 60\% purity on 20NEWS  and $95\%$ purity  MNIST. A dash indicates that the algorithm never arrived to the target purity.}
\end{table}
}

{\bf Speed and Accuracy Comparisons:}  In figure \ref{prtvstime} we report the performance of the selected algorithms LSD, NMFR, MTV and INCRES (with parameter ``speed'' set to either $1$ or $5$) algorithms  on the 20NEWS and MNIST data sets. We quantify performance in terms of both accuracy and speed. We use cluster purity to quantify the accuracy of a calculated partition, defined according to the relation:
$ {\rm Purity} = \frac{\# \text{number of ``successes''}}{N}= \frac{1}{N}\sum_{r=1}^R  \max_{1<i<R} n_{r,i}.$
Here $n_{r,i}$ denotes the number of data points in the $r^{ {\rm th}}$ cluster that belong to the $i^{ {\rm th} }$ ground-truth class. In other words, given a computed cluster we count a data point as a success if it belongs to the ground truth class that best represents the cluster. Each algorithm was run until convergence. Both INCRES and MTV are randomized algorithms, so we obtain their corresponding purity curves by averaging the results over 120 runs. 
In order to give an indication of the speed/accuracy trade-off for each algorithm, in table \ref{tab:crosstime} we record the time it took for each algorithm to reach 60\% purity on 20NEWS and 95\% purity on MNIST.  

 Overall, the simple INCRES algorithm provides accuracy comparable to the state-of-the-art NMF algorithm \cite{yang2012clustering} yet runs an order of magnitude faster. 
 Both algorithms utilize a random walk strategy, which helps to smooth irregular graphs such as the similarity matrices obtained from text data sets. This strategy contributes to the robustness of these algorithms and to their solid performance on a range of datasets. Due to the similarity of their results, we provide a more exhaustive comparison between these two algorithms in the appendix.

 {\bf Accuracy Comparisons:}  Table \ref{tab:prt} reports the cluster purity obtained by each algorithm on all of the test data sets.  We allowed each iterative algorithm a total of 10,000 iterations.  For the randomized algorithms (INCRES and MTV) we report the average purity achieved over 120 different runs. The second column indicates the size  of each of the data sets ($N$) and the number of classes ($R$). The fourth column provides a base-line purity for reference, i.e. the purity obtained by assigning each data point to a class from $1$ to $R$ uniformly at random. \\

{\small
\begin{table}[h!]
\centering 
\begin{tabular}{l|l|l|l|l|l|l|l|l}
{\bf Data} &   $N/R$ &{\bf RND} &  {\bf NCUT} & {\bf LSD}   &{\bf NMFR}& {\bf MTV} & {\bf INCRES}& {\bf INCRES} \\
                       &                  &                      &                      &                    &                    &                  &         {\it (speed 1)}                          &           {\it (speed 5)      }                  \\  
	
20NEWS &20K/20  & 6.3\%&  26.6\% &  34.3\% &         {60.7\%} & 35.8\% & {\bf 61.1\%}  & {60.7\%}  \\
 CADE &21K/3 & 15.5\%  & 41.0\% & 41.3\% & 52.0\% & 44.2\%& {\bf52.9\%} & {52.1\%}\\
 RCV1 &9.6K/4  & 30.3\% & 38.2\% & 38.1\% & 42.7 \%&42.8\% & {\bf 54.6\%} & {51.1\%} \\
WEBKB4 &4.2K/4  & 39.1\% & 39.8\% & 45.8\%   & {\bf 58.06\%}&45.2\%  & {57.0\%} & 56.8\%  \\
CITESEER& 3.3K/6 & 21.8\% & 23.4\% & 53.4\% &  {\bf 62.6 \%}&  42.6\%  &  61.9\% & {62.2\%}  \\
  
MNIST & 70K/10  & 11.3\% & 76.9\% & 75.5\% &  {\bf 97.1\%}& 95.5\% & {96.5\%} & 96.23\%  \\
PENDIG. &  11K/10 & 11.6\% & 80.2\% & 86.1\% & {86.8\%}& 86.5\%&  {\bf88.8\%} & 85.54\%\\
USPS & 9.3K/10 & 16.7\% & 71.5\% &  70.4\% & 86.4\% &85.3\% & {\bf 87.43\%} & {86.7\%} \\
OPTDIG.  & 5.6K/10& 12.0\% & 90.8\%  & 91\% & {\bf 98.0\%}& 95.2\% & {97.2\%} & 95.0\% \\
\end{tabular}
\label{tab:prt}
\caption{Algorithmic comparison via cluster purity. Boldface indicates the highest purity score for each data set.}
\end{table}
}

Once again, these experiments show that the  INCRES algorithm provides accuracy comparable to the NMFR algorithm \cite{yang2012clustering}.  The timing results for these data sets are consistent with those obtained for 20NEWS and MNIST (c.f. figure  \ref{prtvstime} and table \ref{tab:crosstime}), in the sense that INCRES typically runs one order of magnitude faster than NMFR on these data sets as well.

\subsection{Robustness to perturbation} 
\label{sec:penpert}
We took the PENDIGITS matrix from  \cite{yang2012clustering} and uniformly at random added
noise edges.  The original graph had $e=149, 652$ edges; in the experiment, we add $.5e$, $e$, and $2e$ noise edges.  The results are in Table \ref{tab:pert}.  We average results over 16 trials.  Each iterative algorithm was run for 10,000 iterations.

{\small
\begin{table}[h!]
\centering
\begin{tabular}{c|c|c|c|c|c|c|c|c|c}
{\bf Dataset} &   {\bf NCUT}  &{\bf NMFR}& {\bf MTV} & {\bf INCRES}   & {\bf INCRES}  \\
              &               &          &           & {\it (speed 1)}&{\it (speed 5)}\\  
\hline	

PENDIGITS .5&  70.3\% & 84.6\% & 44.6\%&  {84.7\%} &  77.8\%\\
PENDIGITS  1&  64.5\% & 78.7\% & 27.8\% &  {83.2\%} &  76.2\%\\
PENDIGITS  2&  50.7\% & 68.1\% & 16.6\% &  {80.5\%} &  71.0\%\\
\end{tabular}
\label{tab:pert}
\caption{Robustness to perturbation; PENDIGITS $\alpha$ has fraction $\alpha$ noise edges.}
\end{table}
}
\vspace{-.3 in}
\subsection{LFR benchmark}
\label{sec:lfr}
Lancichinetti, Fortunato, and Radicchi have introduced \cite{LFR} a well-known class of synthetic benchmark graphs (the LFR benchmarks) to provide a testbed for community-detection algorithms. 
 Each node in the graph shares a fraction $1-\mu$ of its edges with nodes in its own community and a fraction $\mu$ of its edges with nodes in other communities. The quantity $\mu$ is called the \emph{mixing parameter}, and it provides a measure of how well-defined the communities are in the graph.  If $\mu > 0.5$ then each node shares more than half of its edges with nodes in other communities, and so the communities become increasingly hard to detect past this point. 
  The code used to generate the LFR data is publicly available provided by the authors in \cite{LFR}. 
   In our experiments, we consider a graph with  10,000 nodes, divided in 10 communities of equal sizes. The degree of the nodes are set to 16. We study the behavior of the various algorithms as the mixing parameter varies from 0.45 to 0.65.   The table is shows results averaged over 16 constructions of the data for each mixing parameter. 
%
%
%
%
{\scriptsize
\begin{table}[h!]
\centering
\begin{tabular}{c|c|c|c|c|c|c|c|c}
{\bf mixing parameter} &  {\bf NCUT}    &{\bf NMFR}& {\bf MTV} & {\bf INCRES}      &      {\bf INCRES} \\
                       &                &           &          &  {\it (speed 1)}   &  {\it (speed 5)      }   \\  
\hline	
0.45   & 		     		100\%	    &        100\%        &         45.3\%       &            100\%  &100\%  \\
  0.50 & 		    		99.9\%	   &         100\%      &          28.5\%         &             100\%& 100 \%  \\
0.55    & 		    		96.5\%	   &         99.9\%         &        20.1\%         &             99.4\%& 99.8\%   \\
0.60   & 		    		35.9\%	   &        86.7\%        &          15.3\%         &             88.7\%&  55.7\%  \\
0.65   & 		    		14.8\%	   &        14.7\%        &          13.3\%         &             13.1\%&  13.2\% \\
\end{tabular}
\label{tab:lfr}
\caption{Purity on LFR benchmark datasets for various mixing parameters}
\end{table}
}

\section{Multigrid Experiments}
{\small
\begin{table}[h!]
\centering
\begin{tabular}{l|l|l|l|l|l|l}
{\bf Data} &{\bf size} & {\bf METIS }& {\bf GRACLUS} & {\bf INCRES} &  {\bf INCRES} & {\bf INCRES  } \\
                      &                                      &                    &                                 &       $N_{ {\rm sm} } = 500 $                  &       $N_{ {\rm sm} } = 1500$                &   no refinement     \\
&      &         &                        &  $k_{ {\rm sm} } = 250$  &  $k_{ {\rm sm} } = 125$                     &              \\           
\hline
20NEWS     & 20K  &    42.4\%  & 42.4\% (0.05s)&  {\bf 57.5\%} (1.5s)  & 54.4\% (1.1s) & 36.5\% (0.5s) \\       
CADE       & 21K  &   29.3\%  & {\bf 43.5\%} (0.1s) & 41.8\% (0.9s) & 45.1\% (1.1s) &   40.6\% (0.4s) \\
RCV1       & 9.6K &  34.1\% &  42.4 (0.01s) &  44.0\% (0.3s)&     {\bf 45.2\%}(0.3s) & 42.4\% (0.1s)\\
WEBKB4     & 4.2K &   37.9\%  & 49.0\%(0.01s) & 51.0\%(0.2s)  &{\bf  52.6\%}(0.2s) & 46.7\% (0.1s)  \\
CITESEER   & 3.3K &   45.2\%  &53.5\%(0.01s)&    60.2\%(0.3s)& {\bf 61.6\%}(0.2s) & 54.3\% (0.2s)\\
\hline
MNIST      &  70K &   86.0\%  &   {\bf 96.9\%} (0.17s)  & 96.2\%(3s) &92.7\%(3.2s) & 89\% (0.7s)   \\   
PENDIG.  & 11K  &   67.3\%  &  84.7\% (0.02s) &  {\bf 87.8\%} (0.9s) &  83.4\% (1.1s) & 86\% (0.3s) \\
USPS       & 9.3K &   75.1\%  &   {\bf 86.9\%} (0.02s) & 86.5\%(0.7s)&  86.2\%(0.7s) & 83.9\% (0.3s)\\
OPTIDIG. & 5.6K &   83.0\%  & {\bf 94.2\%} (0.01s)  & 93.0\% (0.6s) & 91.1\% (0.5s) & 92.4\% (0.3s)
 
\end{tabular}
\label{tab:Multigrid}
\caption{Accuracy comparison for multilevel algorithms. All results are averages over 500 trials}
\end{table}
}
Table \ref{tab:Multigrid} shows the accuracies and run times of the coarsen and refine algorithms. Note that the rightmost column, INCRES with no refinement, uses the same algorithm for coarsening as METIS and GRACLUS and only the trivial ``refinement'' to get back to the original graph. Since the coarsened graph is quite small, the only difference in timing between the methods is the {\it implementation} of the coarsening.  

We have found that the coarsen and refine procedure can be very sensitive to inpure neighborhoods.  In particular, these algorithms do very poorly on the benchmarks in Section \ref{sec:penpert} and \ref{sec:lfr}.


\section*{Appendix}

\small{
\begin{table}[h!]
\centering
\begin{tabular}{c|c|c|c|c|c|c|}
{\bf Dataset} &{\bf size}& {\bf RAND} & {\bf NMFR } & {\bf INCRES }  \\ 
	             &                 &                      &                                        &   (speed  1)  \\   
\hline
%
%
%
%

YEAST  & 1.5K & 32\% & {\bf 55}\% & 54\% \\  
SEMEION & 1.6K & 13\% & {\bf 94}\% & 93\% \\ 
FAULTS  & 1.9K & 34\% & 39\% & {\bf 41}\%\\   
SEG &2.3K & 16\% & {\bf 73}\% & 59\% \\  

CORA  & 2.7K & 30\% & {\bf 47}\% & 46\%  \\ 
MIREX & 3.1K & 13\% & 43\% & {\bf 45}\% \\  
CITESEER & 3.3K & 21\% & 44\% & {\bf 47}\% \\ 
WEBKB4 & 4.2K & 39\% & {\bf 63}\% & 60\%\\   
7SECTORS & 4.6K & 24\% &  {\bf 34}\% & {32}\% \\  

SPAM  & 4.6K & 60\% & {\bf 69}\% & { {\bf 69}\%} \\ 
CURETGREY &5.6K & 5\% & {\bf 28}\% & {16}\% \\ 
OPTDIGITS &5.6K & 12\% & {\bf 98}\% & {\bf 98}\%\\  
GISETTE &7.0K &50\% & {\bf 94}\% &{\bf  94}\% \\ 
REUTERS &8.3K & 45\% & {\bf 77}\% & {74}\%\\  

RCV1  &9.6K & 30\% & 54\% &  {{\bf 56}\%}  \\  
PENDIGITS &11K& 12\% & 87\% &  {{\bf 89}\%} \\ 
PROTEIN & 18K & 46\% & {\bf 50}\% &  {46\%}  \\ 
20NEWS &20K & 6\% & {\bf 63}\% & {{\bf 63}\%} \\ 
MNIST & 70K & 11\% & {\bf 97}\% & {96}\% \\ 
SEISMIC &99K & 50\% & {\bf 59}\% & {56}\%  

\end{tabular}
\caption{Algorithmic comparison via cluster purity. Boldface indicates the highest purity score for each data set.}
\end{table}

}

The table above provides a more exhaustive comparison between the INCRES and NMFR algorithms. We selected the twenty largest data sets   used in the original NMFR paper  \cite{yang2012clustering}.  We excluded the  ADS dataset because the similarity matrix contained negative entries and no algorithm performed better than random on this data set. The similarity matrices  were downloaded  from  \url{http://users.ics.aalto.fi/rozyang/nmfr/index.shtml}.  
The similarity matrices for the  text data sets 20NEWS, RCV1 and WEBKB4 are different than the one presented in the main body of our paper. The authors of the original NMFR paper used the 10,000 words with maximum information gain to construct the similarity matrices associated to these text datasets. We have preferred to simply use the words appearing more than  a certain number of times to construct our similarity matrices (in order to avoid using ground truth information in the construction of the similarity matrices). For the NMFR algorithm the results  included in the above table are the one  reported in \cite{yang2012clustering}.

On most of these datasets INCRES and NMFR obtain clustering of similar quality.  NMFR tend to be a little more accurate and consistent, but at the cost of being  one order of magnitude slower.

\subsection*{Datasets}
\begin{itemize}
\item 20NEWS (unweighted similarity matrix):  The word count  from the raw documents was computed using the  Rainbow library \cite{McCallumLibbow} with a default list of stop words. Words appearing less than 20 times were also removed. The similarity matrix was then obtained by 5 nearest neighbors using cosine similarity between tf-idf features.
Source: \url{http://www.cs.cmu.edu/~mccallum/bow/rainbow/}
\item CADE (unweighted similarity matrix): The documents in the Cade12 data set correspond to a subset of web pages extracted from the CAD� Web Directory, which points to Brazilian web pages classified by human experts.   We only kept the three largest classes. The word count  from the raw documents was computed with the Rainbow library \cite{McCallumLibbow} as before. Words appearing less than 5 times were removed. The similarity matrix was then obtained by 5 nearest neighbors using cosine similarity between tf-idf features.
 \\
Source: \url{http://web.ist.utl.pt/\%7Eacardoso/datasets/}.

\item RCV1 (weighted similarity matrix): This  dataset  was obtained in preprocessed format from \url{http://www.cad.zju.edu.cn/home/dengcai/Data/TextData.html}  with the tf-idf features were already computed. We then simply used cosine similarity and 5-NN.

\item   WEBKB4 (unweighted similarity matrix): The word count  from the raw documents was done with the  Rainbow library \cite{McCallumLibbow}.  A list of stop word was removed. Words appearing less than 5 times were removed. The similarity matrix was then obtained by 5 nearest neighbors using cosine similarity between tf-idf features.
Source: \url{http://www.cs.cmu.edu/afs/cs/project/theo-20/www/data/}

\item  CITESEER (weighted similarity matrix):  This  dataset  was obtained in preprocessed format from  \url{http://linqs.cs.umd.edu/projects//projects/lbc/index.html} where  each publication in the dataset is described by a 0/1-valued word vector indicating the absence/presence of the corresponding word from the dictionary.  We then simply used cosine similarity and 5-NN.

\item MNIST, PENDIGITS, OPTDIGITS (unweighted similarity matrix):   The similarity matrices were obtained from \cite{yang2012clustering}, where the authors  first extracted scattering features  using \cite{mallat} for images before calculating the 10-NN graph. 
Source: \url{http://users.ics.aalto.fi/rozyang/nmfr/index.shtml}

\item USPS (weighted similarity matrix): We computed a 10-NN graph using standard Euclidean distance between the raw images. Each edge in the 10-NN graph was given the weight
$$
w_{ij} = \mathrm{e}^{ - \frac{\|\mathbf{x}_i - \mathbf{x}_{j}\|^{2}}{2\sigma^{2} } }
$$
where each $\mathbf{x}_i$ denotes a vector containing the raw pixel data. The parameter $\sigma$ was chosen as the mean distance between each vertex and its $10^{ {\rm th }}$ nearest neighbor. Source: \url{http://www.cad.zju.edu.cn/home/dengcai/Data/MLData.html}

\end{itemize}

\bibliographystyle{unsrt}
\bibliography{bib_nips}

\end{document}